\documentclass[10pt,a4paper]{article}
\usepackage[utf8]{inputenc}
\usepackage{amsmath}
\usepackage{amsfonts}
\usepackage{amssymb}
\usepackage{physics}
\usepackage{graphicx}
\usepackage{multirow}
\usepackage{algorithm}
\usepackage{algorithmic}
\usepackage{threeparttable}

\begin{document}
\title{Discriminative multi-view Privileged Information learning for image re-ranking}
\author{Jun Li, Chang Xu, Wankou Yang, Changyin Sun, Dacheng Tao, \\ Hong Zhang
\thanks{Jun Li, Wankou Yang and Changyin Sun are with the School of Automation, Southeast University, Nanjing 210096, China (email lijunautomation@gmail.com;
youngwankou@yeah.net; cysun@seu.edu.cn).
Chang Xu and Dacheng Tao are with the UBTech Sydney AI Institute, The School of
Information Technologies, The University of Sydney, Darlington, NSW 2008,
Australia (e-mail: c.xu@sydney.edu.au; dacheng.tao@sydney.edu.au).
Hong Zhang is with the Department of Computing Science, University of Alberta, Edmonton, AB T6G 2E8, Canada (email: hzhang@ualberta.ca).}
}
\date{\today}
\maketitle

\begin{abstract}
Conventional multi-view re-ranking methods usually perform asymmetrical matching between the region of interest (ROI) in the query image and the whole target image for similarity computation. Due to the inconsistency in the visual appearance, this practice tends to degrade the retrieval accuracy particularly when the image ROI, which is usually interpreted as the image objectness, accounts for a smaller region in the image. Since Privileged Information (PI), which can be viewed as the image prior, enables well characterizing the image objectness, we are aiming at leveraging PI for further improving the performance of the multi-view re-ranking accuracy in this paper. Towards this end, we propose a discriminative multi-view re-ranking approach in which both the original global image visual contents and the local auxiliary PI features are simultaneously integrated into a unified training framework for generating the latent subspaces with sufficient discriminating power. For the on-the-fly re-ranking, since the multi-view PI features are unavailable, we only project the original multi-view image representations onto the latent subspace, and thus the re-ranking can be achieved by computing and sorting the distances from the multi-view embeddings to the separating hyperplane. Extensive experimental evaluations on the two public benchmarks Oxford5k and Paris6k reveal our approach provides further performance boost for accurate image re-ranking, whilst the comparative study demonstrates the advantage of our method against other multi-view re-ranking methods.
\end{abstract}

\section{Introduction} \label{section1} 
Recent years have witnessed massive efforts devoted to advancing the research over image re-ranking which allows significantly improving the retrieval accuracy by refining the query model. Among all the re-ranking approaches, the subspace based strategy has become a promising line of research due to its desirable property in uncovering the discriminative latent subspace underlying the original high-dimensional feature space. In particular, multi-view re-ranking methods are capable of exploring the visual complementarity among heterogeneous feature spaces, which, therefore, leads to a latent representation with sufficient descriptive power. In order to further improve the separability of the query model, the classification mechanism is usually encoded into the subspace learning based re-ranking method for producing a generic and discriminative framework\cite{JunTIP17,ArandjelovicCVPR12}.

Despite their success in image re-ranking, conventional subspace based approaches directly leverage the visual features generated from the whole image for training the query model while ignores the important role of image objectness in similarity matching. In many cases, actually, it is the region of interest (ROI) characterizing the image objectness that captures the users' query intention rather than the whole image region containing complex background contents. In this sense, training the query model without considering the objectness tends to introduce the query-irrelevant noise, which leads to the biased re-ranking results and thus adversely affects the retrieval performance. Therefore, it is crucial for incorporating the objectness into the trained query model for further improving the re-ranking accuracy. 

It is well known that Privileged Information (PI) gives the supplementary cues about the training examples \cite{VapnikNN09,SharmanskaICCV13}. Since PI is typically more informative about the task at hand than the raw data per se, it is usually combined with the original training examples for further improving the accuracy of the trained model. Recent research substantially demonstrates the beneficial effect of PI learning in a wide range of vision tasks. Without loss of generality, PI can be defined as four different modalities in the context of object classification, namely attributes, annotator rationales, bounding boxes and textual descriptions \cite{SharmanskaICCV13}. In particular, the PI translated into the bounding boxes can be viewed as the image prior, since it is capable of highlighting the object region and encoding the principal visual cues in the image. Besides, it is also available with easy-to-implement ROI annotation, which is tailored for the user interaction in image re-ranking. Therefore, in this paper, we only focus on the PI formulated as the bounding box, whilst aim to exploit both the original and the supplementary PI features to train the re-ranking model for accurate retrieval. More specifically, inspired by the unified subspace based re-ranking framework proposed in \cite{JunTIP17}, we propose a discriminative PI-aware multi-view re-ranking method in which multi-view local PI features are also integrated into the query model training along with their global counterparts.

\begin{figure}
\centering
\includegraphics[width=1.0\linewidth]{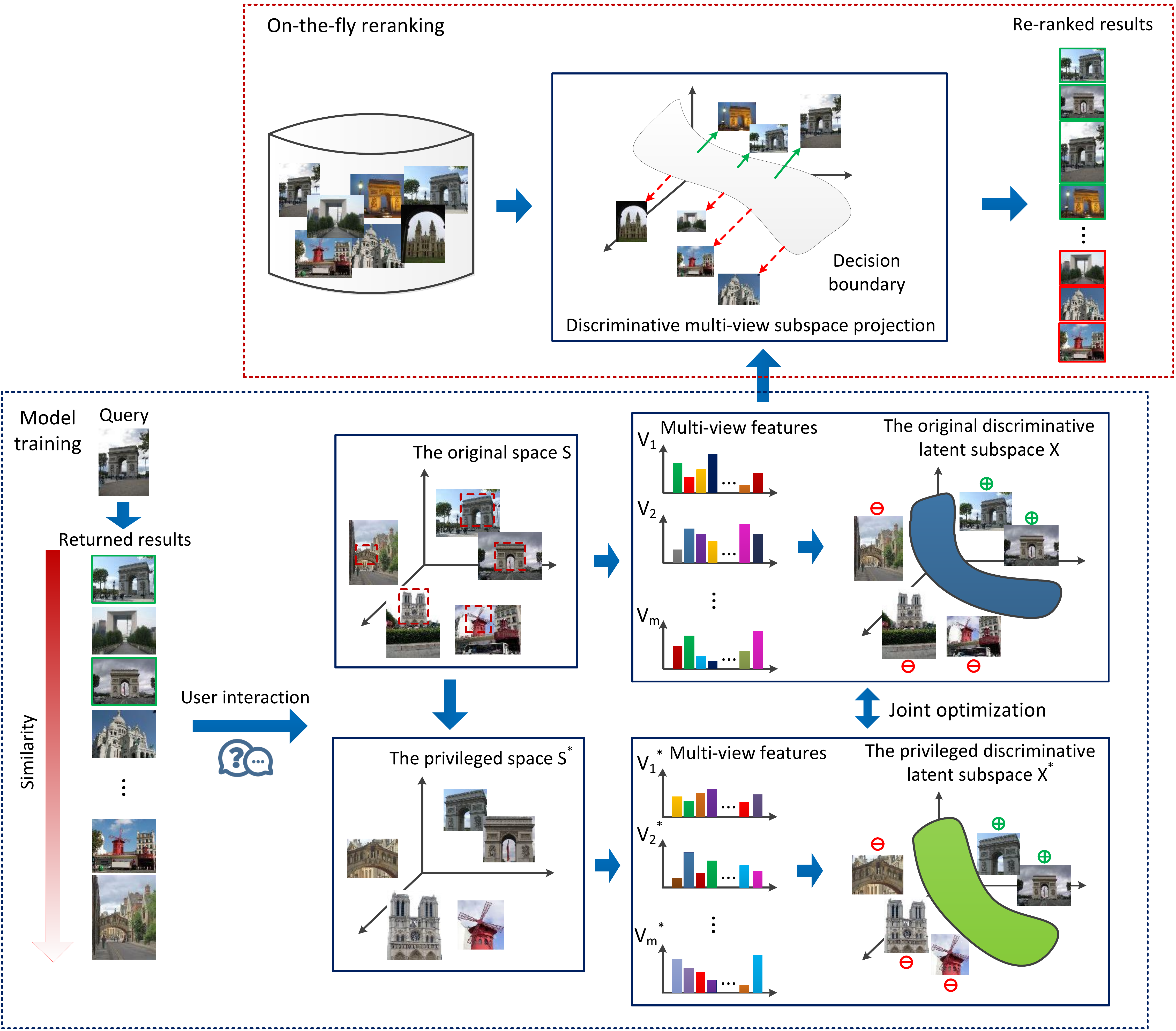}
\caption{The system flowchart of the proposed approach. In our method, two major steps are involved, namely model training and on-the-fly re-ranking. With the original training data and the annotated PI regions derived from user interaction, the former integrates the global multi-view embedding and its PI-based counterpart into a unified framework, which produces a PI-aware latent subspace with sufficient discriminating power. For on-the-fly re-ranking, the latent representations of the target images are obtained by projecting the multi-view features onto the PI-aware subspace, and thus the images are re-ranked by the signed distances from the separating hyperplane accordingly.}
\label{fig1} 
\end{figure}

Fig. \ref{fig1} gives the processing pipeline of the proposed method. Analogous to the DMINTIR re-ranking method in \cite{JunTIP17}, our approach comprises two steps, namely query model training and on-the-fly re-ranking. In the model training, we first identify the query-relevant images from the top returned shortlist, whilst annotate the PI regions in these positive examples with cropped ROI bounding boxes via user interaction. In addition, the low-scored images in the original ranking list are automatically recognized as the negative training examples, and their corresponding PI regions can be obtained by the off-the-shelf saliency detector \cite{FengICCV11}. Thus, the training data consisting of both global contents and additional local PI regions can be handled in the original and the privileged spaces respectively. Then, we compute the multi-view features in both spaces and project them onto the respective latent subspaces for uncovering the underlying low-dimensional representations. Meanwhile, a PI-aware latent subspace with sufficient discriminating power can be obtained by jointly optimizing the separating hyperplanes of the dual subspaces. For the on-the-fly re-ranking, due to the unavailability of the PI in the target images, we directly project the multi-view features onto the PI-aware latent subspace for generating the discriminative representations, and thus the database images can be re-ranked by computing and sorting the distances from the separating hyperplane for performance improvement.

\begin{figure}  
\centering
\includegraphics[width=1.0\linewidth]{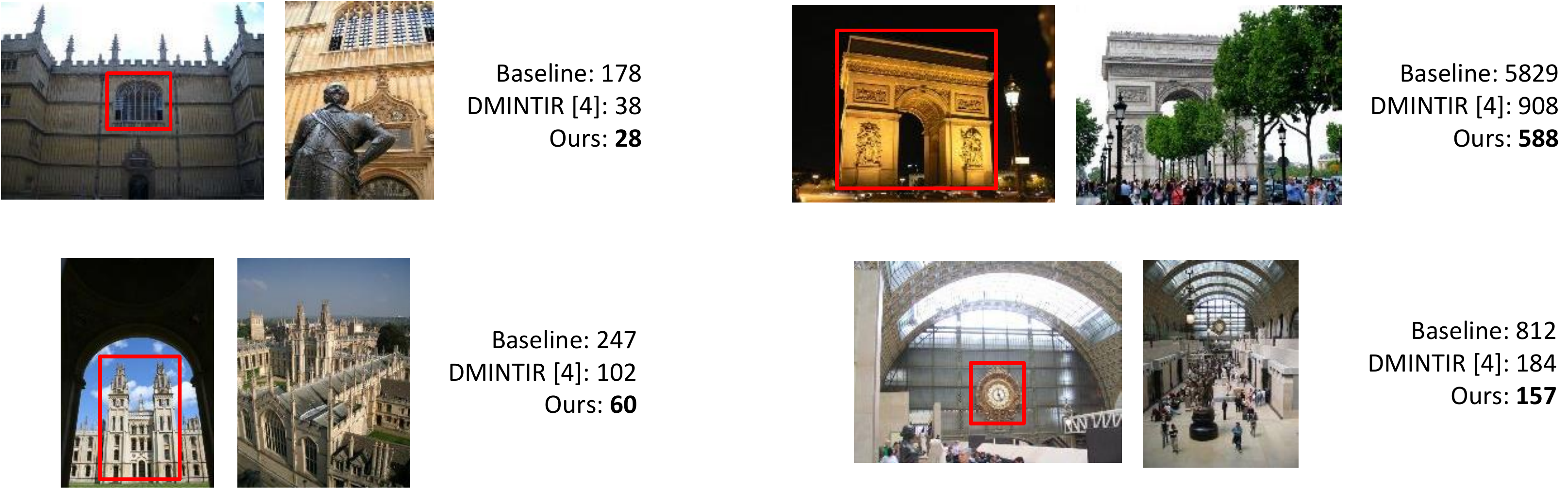}
\caption{Some difficult query examples on Oxford5k and Paris6k. The left image with a red box is the query with the annotated image ROI while the right image is the corresponding target image in the dataset. The numbers beside the target images denote the image ranks in the returned list. Note that the DMINTIR proposed in \cite{JunTIP17} is closest to our method except excluding the PI features in training the query model. Apparently, our method significantly outperforms the DMINTIR when there exists dramatic variances in visual appearances of query object and background contents, which sufficiently suggests the benefits of PI for accurate re-ranking.}
\label{fig2}
\end{figure}

Fig. \ref{fig2} presents some difficult query examples by using different approaches. It is shown that there exists dramatic visual variances in ROI regions and background contents between the query images and the target images. Since our approach takes into account the PI cues in training re-ranking model, it outperforms the state-of-the-art re-ranking method \cite{JunTIP17} which excludes PI features in model training. This also sufficiently suggests the beneficial effect of PI for improving the retrieval accuracy. Specifically, the advantage of exploiting PI for re-ranking manifests itself in the following two aspects. On the one hand, training with PI contributes to highlighting the dominant role of query object in the re-ranking model for further enhancing its discriminating power. On the other hand, PI-aware training somewhat allows suppressing the adverse effect of scale variance, illumination change, perspective transformation and cluttered background, and thus improves the robustness of the re-ranking model.

To sum up, the contributions of this paper are three-fold as follows:

\begin{itemize}
\item We take into account the PI clues in training re-ranking to outweigh the objectness in the image. To our knowledge, this is the first time PI is involved in re-ranking for further performance boost.

\item We simultaneously integrate the local PI features and the original global features into a unified PI-aware multi-view embedding framework for accurate image re-ranking.

\item Extensive experimental evaluations and the comparative studies demonstrate the advantage of our method to other state-of-the-art approaches.
\end{itemize}

The remainder of this paper is structured as follows. After reviewing the related work in Section \ref{section2}, we will introduce the problem setup in Section \ref{section3}. Subsequently, we will elaborate the mathematical formulation of our method in Section \ref{section4} and give the optimization accordingly in Section \ref{section5}. Next, we will analyze the computational complexity in Section \ref{section6}. Extensive experimental evaluations and the comparative studies will be carried out in Section \ref{section7} before this paper is finally concluded in Section \ref{section8}.

\section{Related Work} \label{section2} 

\subsection{Multi-view image re-ranking}
In re-ranking, the single-view feature often fails to provide a comprehensive visual description, and thus leads to an image signature with insufficient descriptive power. By contrast, multi-view feature enables take advantage of the complementarity among multiple heterogeneous features, which, consequently, substantially benefits the re-ranking performance improvements. Earlier multi-view re-ranking approaches leverage the low-level features (e.g., Bag of features, color histogram and wavelet textures) for characterizing the visual contents in the images \cite{YuTC15,YuTIP14,YuTMM14,DengTMM14,LiTMM12}. Then, either a linear transformation \cite{LiTMM12} or complex hypergraph manifolds \cite{YuTC15,YuTIP14,YuTMM14} are learned from these multi-view features to uncover the intrinsic structure or a low-dimensional subspace for re-ranking. Besides, more robust estimator has also been utilized for multi-view intact space learning \cite{XuTPAMI15}.

Low-level features encodes the visual patterns intuitively yet fails to provide higher-level image representation. Recently, deep features have been used as desirable alternative in multi-view learning, since they encode high-level semantic attributes in the image with preferable descriptive power \cite{JiINS17,JunTIP17,PanSIGIR14}. Particularly, a discriminative multi-view re-ranking approach has been proposed in \cite{JunTIP17} to integrate the deep CNN code and the best performing hand-crafted feature TE into a generic and unified framework, which produces a latent low-dimensional subspace maintaining sufficient separability. Thus, multi-view features can be projected onto this subspace such that robust latent representations can be generated for accurate re-ranking. Albeit effective, \cite{JunTIP17} directly exploits the global features for multi-view embedding while downplays visual cues in the query region. Therefore, it exhibits suboptimal performance when there exists complex background contents and severe geometric transformation of query object.

\subsection{Learning using privileged information}
In the computer vision community, PI, which is interpreted as the auxiliary information about the training data, can be used for learning better recognition systems. Recently, extensive efforts are devoted to exploring PI cues for enhancing the model training in a variety of vision tasks \cite{SharmanskaICCV13,XuTNNLS15,YanTMM16,MotiianCVPR16,YangCVPR17,ShiCVPR17,GordoCVPR17,TangTNNLS17,YangTIP18,LambertCVPR18}. The earliest research over PI learning integrates PI into the classic SVM algorithm, which produces an extended paradigm termed Learning Using Privileged Information (LUPI) \cite{VapnikNN09}. The resulting model is also referred to SVM+ accordingly. In \cite{SharmanskaICCV13}, four different PI types are explored and handled in a unified LUPI framework in the context of object classification. Besides, a novel rank transfer approach comparable to the conventional SVM+ algorithm is also proposed for solving the LUPI task. While the extensions of SVM+ algorithm to multiclass problem are possible, PI is also incorporated into the framework of generalized matrix learning vector for prototype-based classification \cite{FouadTNNLS13}. In face verification and person re-identification, the extra depth features used as PI are utilized for improving the distance metric learning \cite{XuTNNLS15}. Analogously, person re-identification is also addressed in \cite{YangTIP18} by joint distance metric learning with the help of PI. Besides, PI is also embedded into the deep Convolutional Neural Networks (CNNs) and Recurrent Neural Network (RNNs) for image classification and action recognition \cite{LambertCVPR18,YangCVPR17,ShiCVPR17}. In addition to the aforementioned applications, human-generated captions is used as PI for learning the improved representation in semantic retrieval \cite{GordoCVPR17}. In order to further exploit the complementary information among multiple features sets, a new multi-view privileged SVM model is proposed by incorporating the LUPI paradigm into multi-view learning framework, which satisfies both consensus and complementary principles for multi-view learning \cite{TangTNNLS17}.

Although great progress has been made in PI learning, how to make use of PI cues in image re-ranking for further performance improvement remains an open problem. In this paper, we propose a generic PI-aware re-ranking framework in which the original global representations and the additional PI cues are simultaneously incorporated into subspace-based multi-view embedding. The resulting PI-aware subspace preserves sufficient discrimination in the image, and thus can be used for generating discriminative objectness-aware latent representation for accurate re-ranking. To the best of our knowledge, this is the first time the PI learning is explored in image re-ranking.

\section{The problem formulation} \label{section3} 
Given a set of training examples $ \mathcal{S}=\{S^+,S^-\} $ with annotated ROI bounding boxes obtained by user interaction, the corresponding multi-view features generated from both the whole image and ROI can be denoted as $ \mathcal{Z}=\{Z_v\}_{v=1}^m $ in the original space and $ \mathcal{Z}^*=\{Z_v^*\}_{v=1}^m $ in the privileged space, where $ m $ is the number of views. Since both spaces share the same feature dimensionality for single-view data, we have $ Z_v, Z_v^* \in \mathbb{R}^{D_v \times n} $, where $ D_v $ is the feature dimensionality of the $ v^{th} $ view while $ n=|\mathcal{S}| $ is the size of the training set. Meanwhile, $ \mathcal{Y} \in \{+1,-1\}^{n \times 1} $ is the label vector denoting the query relevance of the training examples. The PI-aware multi-view re-ranking model training is aiming at learning the dual mapping functions:
\begin{align}
f&:\{\mathcal{Z},\mathcal{Y}\} \rightarrow \{X,w\} \label{equ1} \\
f^*&:\{\mathcal{Z^*},\mathcal{Y}\} \rightarrow \{X^*,w^*\} \label{equ2} 
\end{align}
where $ X $ and $ X^* $ are the respective multi-view subspace embeddings, whilst $ w $ and $ w^* $ are the dual separating hyperplanes preserving sufficient discriminative power in both underlying subspaces. For the sake of consistency, we learn the function $ h(w,w^*) $ such that the two subspaces are mutually interlinked and a PI-aware low-dimensional subspace can be produced. 

For the on-the-fly re-ranking, since the PI data is unavailable, we directly project the multiple features of the target images $ \{\tilde{Z}_v\}_{v=1}^m $ on the trained PI-aware subspace for generating the latent representations $ \tilde{X} $. Thus, the image ranks can be refined by computing the signed distance of $ \tilde{X} $ from the decision boundary $ w $ for accurate re-ranking. All the mathematical notations involved in our formulation are summarized in Table \ref{tab1}.

\begin{table} \addtolength{\tabcolsep}{+0.1cm} \renewcommand{\arraystretch}{1.2} 
\centering
\caption{The annotations involved in our mathematical formulation}
\label{tab1} 
\vspace{+.1cm}
\begin{tabular}{|c|c|}
\hline 
$n$ & the size of the training set \\ 
\hline 
$m$ & the number of data view \\ 
\hline
$D_v$ & the view-specific feature dimensionality \\
\hline
$d$ & the dimension of the latent subspace \\
\hline 
$z_v^{(i)} \in \mathbb{R}^{D_v \times 1}$ & the view-specific feature representation in the original space \\
\hline 
$z_v^{*(i)} \in \mathbb{R}^{D_v \times 1}$ & the view-specific feature representation in the privileged space \\
\hline 
$P_v \in \mathbb{R}^{D_v \times d}$ & the view-specific generation matrix in the original space \\
\hline 
$P_v^* \in \mathbb{R}^{D_v \times d}$ & the view-specific generation matrix in the privileged space \\
\hline 
$x_i \in \mathbb{R}^{d \times 1}$ & the sample-specific latent representation in the original space \\ 
\hline 
$x_i^* \in \mathbb{R}^{d \times 1}$ & the sample-specific latent representation in the privileged space \\ 
\hline
$w \in \mathbb{R}^{d \times 1}$ & the separating hyperplane in the original space \\ 
\hline 
$w^* \in \mathbb{R}^{d \times 1}$ & the separating hyperplane in the privileged space \\ 
\hline 
$y_i \in \{1, -1\}$ & the sample label \\ 
\hline 
\end{tabular}
\end{table}

\section{DMVPIR: Discriminative Multi-view PI-aware Re-ranking} \label{section4}

In the state-of-the-art subspace-based multi-view embedding methods, it is assumed that the image feature of a single view $ z_v \in \mathbb{R}^{D_v \times 1} $ can be recovered from a shared underlying subspace via a view-specific generation matrix $ P_v \in \mathbb{R}^{D_v \times d} $ such that:
\begin{equation} \label{equ3} 
z_v=P_v \cdot x + \epsilon_v
\end{equation}
where $ x \in \mathbb{R}^{d \times 1} $ is the low-dimensional subspace representation, whilst $ \epsilon_v $ is the view-dependent mapping error. Thus, the latent subspace can be obtained by minimizing the following formulation:
\begin{equation} \label{equ4} 
J(P_v,x)=\sum_{v=1}^m\|z_v-P_v \cdot x\|^2 + \lambda\sum_{v=1}^m\|P_v\|_F^2 + \beta\|x\|^2
\end{equation}
where $ \lambda $ and $ \beta $ are the tradeoff parameters compromising between the two regularization terms.

In our case, we impose the multi-view embedding on all the training examples in both the original and the privileged feature spaces, and thus we have the following formulations to minimize:
\begin{align}
J(P_v,x_i) &= \sum_{i=1}^n\sum_{v=1}^m\|z_v^{(i)}-P_v \cdot x_i\|^2 + \lambda\sum_{v=1}^m\|P_v\|_F^2 + \beta\sum_{i=1}^n\|x_i\|^2 \label{equ5}  \\
J(P_v^*,x_i^*) &= \sum_{i=1}^n\sum_{v=1}^m\|z_v^{*(i)}-P_v^* \cdot x_i^*\|^2 + \lambda^*\sum_{v=1}^m\|P_v^*\|_F^2 + \beta^*\sum_{i=1}^n\|x_i^*\|^2 \label{equ6} 
\end{align}

To ensure the model discrimination capability, learning separating hyperplane $ w $ and $ w^* $ in the dual subspaces should also be also encoded in the formulation to distinguish between query-relevant and irrelevant examples. Besides, $ w^* $ learning should play a dominant and leading role, since the privileged features are more informative and confident in discriminatively separating the examples. As a result, simultaneous learning of $ w $ and $ w^* $ can be formulated as: 
\begin{equation} \label{equ7} 
\begin{gathered}
\min_{w,w^*,b,b^*}\dfrac{1}{2}(\|w\|^2+\gamma\|w^*\|^2) + C\sum_{i=1}^n w^{*T}x_i^*+b^* \\
s.t. \quad y_i(w^Tx_i + b) \geq 1-(w^{*T}x_i^* + b^*), \quad w^{*T}x_i^* + b^* \geq 0,\quad \forall i=1,...,n
\end{gathered} 
\end{equation}

Note that Equ.(7) is actually the formulation of SVM+ algorithm which uses the PI as a proxy to the slack oracle in the traditional SVM classifier \cite{SharmanskaICCV13}. Thus, we have the mathematical formulation of our Discriminative Multi-View PI aware Re-ranking (DMVPIR) model by integrating (5), (6) and (7) into a unified framework as follows:
\begin{equation} \label{equ8}
\begin{gathered}
\mathcal{L}(x_i, x_i^*, P_v, P_v^*, w, w^*) = \min J(P_v, x_i) + J(P_v^*, x_i^*) \\
+ \frac{1}{2}(\|w\|^2+\gamma\|w^*\|^2) + C\sum_{i=1}^n{w^*}^Tx_i^* + b^* \\
s.t.\quad y_i(w^Tx_i + b) \geq 1-(w^{*T}x_i^* + b^*), \quad w^{*T}x_i^* + b^* \geq 0,\quad \forall i=1,...,n 
\end{gathered} 
\end{equation}
where
\begin{align}
J(P_v, x_i) &= \sum_{i=1}^n\sum_{v=1}^m\|z_v^{(i)}-P_vx_i\|^2 + \lambda\sum_{v=1}^m\|P_v\|_F^2 + \beta\sum_{i=1}^n\|x_i\|^2 \label{equ9}  \\
J(P_v^*, x_i^*) &= \sum_{i=1}^n\sum_{v=1}^m\|z_v^{*(i)}-P_v^*x_i^*\|^2 + \lambda^* \sum_{v=1}^m\|P_v^*\|_F^2 + \beta^*\sum_{i=1}^n \|x_i^*\|^2 \label{equ10} 
\end{align}

As shown in Equ. (\ref{equ8}), our DMVPIR re-ranking model aims to learn a PI-aware subspace with sufficient discriminative power encoded by decision boundary $ w $. For the on-the-fly re-ranking, we project the multi-view feature representations of the target images $\{\tilde{Z_v}\}_{v=1}^m$ onto the PI-aware latent subspace via the optimal learned view-dependent generation matrix $\{\hat{P_v}\}_{v=1}^m$, which results in the low-dimensional subspace representations $\tilde{X}$ for the subsequent similarity measure and re-ranking. Mathematically, $\tilde{X}$ can be obtained by solving for the following minimization problem:  
\begin{equation} \label{equ11}
\min_{\tilde{X}}\mathcal{L}(\tilde{X}) = \min_{\tilde{X}}\sum_{v=1}^m\|\tilde{Z_v}-\hat{P_v}\tilde{X}\|_F^2 + \beta\|\tilde{X}\|_F^2
\end{equation}


\section{Optimization} \label{section5} 
To solve the problem in Equ. (\ref{equ8}), we develop an efficient iterative alternating optimization algorithm in which the following five alternating optimization steps iteratively minimize the empirical loss.

\textbf{First, we update $x_i$ by fixing the other parameters}, and thus the problem is reduced to the following formulation:

%

\begin{equation} \label{equ12} 
\begin{gathered}
\min_{x_i}\mathcal{L} = \min_{x_i}\sum_{v=1}^m\|z_v^{(i)}-P_vx_i\|^2 + \beta\|x_i\|^2 \\
s.t. \quad y_iw^Tx_i \geq c
\end{gathered}
\end{equation}
where $c=1-(x_i^{*T}w^*+b^*)-y_ib$.

Furthermore, the objective function in Equ. (\ref{equ12}) can be simplified as:

\begin{equation} \label{equ13} 
\begin{aligned}
\min_{x}\mathcal{L} & = \min_{x}\sum_{v=1}^m\|z_v-P_vx\|^2 + \beta\|x\|^2 \\
& = \min_{x}\sum_{v=1}^m({z_v}^T - x^TP_v^T)(z_v-P_vx) + \beta x^Tx \\
& = \min_{x}\sum_{v=1}^m(x^TP_v^TP_vx - 2z_v^TP_vx) + \beta{x^Tx} \\
& = \min_{x}{x^T}{\sum_{v=1}^mP_v^TP_v}{x} + \beta{x^Tx} - 2\sum_{v=1}^m{z_v^TP_v}x \\
& = \min_{x}x^T(\sum_{v=1}^mP_v^TP_v + \beta\textbf{I})x - (2\sum_{v=1}^m{z_v^TP_v})x
\end{aligned}
\end{equation}

Thus, the problem is formulated as:

\begin{equation} \label{equ14} 
\begin{gathered}
\min_{x}\frac{1}{2}x^T(2\sum_{v=1}^mP_v^TP_v + 2\beta\textbf{I})x + (-2\sum_{v=1}^m{z_v^TP_v})x \\
s.t. \quad yw^Tx \geq c
\end{gathered}
\end{equation}
where $c=1-(x^{*T}w^*+b^*)-yb$.

Note that Equ. (\ref{equ14}) is the classic quadratic programming (QP) problem:

\begin{equation} \label{equ15} 
\begin{gathered}
\min_{x}\frac{1}{2}x^TUx + V^Tx \\
s.t. \quad Ax \leq g
\end{gathered}
\end{equation}
where:
\begin{equation} \label{equ16} 
\begin{gathered}
U = 2\sum_{v=1}^mP_v^TP_v + 2\beta\textbf{I}, \quad V = -2\sum_{v=1}^m{P_v^Tz_v} \\
A = -yw^T, \quad g = -c
\end{gathered}
\end{equation}

Thus, Equ. (\ref{equ15}) can be solved by using the QP solver at hand.


\textbf{Second, we update $x_i^*$ by fixing the other parameters}, and thus the problem is reduced to the following formulation:
%

\begin{equation} \label{equ17} 
\begin{gathered}
\min_{x_i^*}\mathcal{L} = \min_{x_i^*}\sum_{v=1}^m\|z_v^{*(i)}-P_v^*x_i^*\|^2 + \beta^*\|x_i^*\|^2 + Cw^{*T}x_i^* \\
s.t. \quad x_i^{*T}w^* \geq c \\
x_i^{*T}w^* \geq -b^*
\end{gathered}
\end{equation}
where $c=1-y_i(w^Tx_i+b) - b^*$.

For the sake of simplicity, the objective function in Equ. (\ref{equ17}) can be expressed as:
\begin{equation} \label{equ18} 
\begin{aligned}
\min_{x^*}\mathcal{L} & = \min_{x^*}\sum_{v=1}^m\|z_v^*-P_v^*x^*\|^2 + \beta^*\|x^*\|^2 + Cw^{*T}x^* \\
& = \min_{x^*}\sum_{v=1}^m({x^*}^T{P_v^*}^TP_v^*x^* - 2{z_v^*}^TP_v^*x^*) + \beta^*{{x^*}^Tx^*} + C{w^*}^Tx^* \\
& = \min_{x^*}(\sum_{v=1}^m{x^*}^T{P_v^*}^TP_v^*x^* + \beta^*{{x^*}^Tx^*}) - \sum_{v=1}^m2{z_v^*}^TP_v^*x^* + C{w^*}^Tx^* \\
& = \min_{x^*}{x^*}^T(\sum_{v=1}^m{P_v^*}^TP_v^* + \beta^*\textbf{I})x^* + (-\sum_{v=1}^m2{z_v^*}^TP_v^*x^* + C{w^*}^Tx^*)
\end{aligned}
\end{equation}

Thus, the problem can be formulated as:

\begin{equation} \label{equ19} 
\begin{gathered}
\min_{x^*}{x^*}^T(\sum_{v=1}^m{P_v^*}^TP_v^* + \beta^*\textbf{I})x^* + (-\sum_{v=1}^m2{z_v^*}^TP_v^* + C{w^*}^T)x^* \\
s.t. \quad {w^*}^Tx^* \geq c, \quad {w^*}^Tx^* \geq -b^*
\end{gathered}
\end{equation}

Apparently, the problem in Equ. (\ref{equ19}) can be also interpreted as a QP problem formulated as
\begin{equation} \label{equ20} 
\begin{gathered}
\min_{x^*}\frac{1}{2}{x^*}^TUx^* + V^Tx^* \\
s.t. \quad Ax^* \leq g
\end{gathered}
\end{equation}
where:
\begin{equation} \label{equ21} 
\begin{gathered}
U = 2\sum_{v=1}^m{P_v^*}^TP_v^* + 2\beta^*\textbf{I}, \quad V = Cw^* - \sum_{v=1}^m{2{P_v^*}^Tz_v^*} \\
A = -{w^*}^T, \quad g = -max(c, -b^*)
\end{gathered}
\end{equation}

Analogously, the problem in Equ. (\ref{equ20}) can also be solved by an off-the-shelf QP solver.

\textbf{Third, we update $P_v$ by fixing the other parameters}, and thus the problem is reduced to the following formulation:

\begin{equation} \label{equ22} 
\min_{P_v}\mathcal{L} = \min_{P_v}\sum_{v=1}^m\|Z_v-P_vX\|_F^2 + \lambda\|P_v\|_F^2 \\
\end{equation}

Equ. (\ref{equ21}) is a unconstrained ridge regression optimization, which could be transformed into:
\begin{equation} \label{equ23} 
\begin{aligned}
\min_{P_v}\mathcal{L} & = \min_{P_v}\sum_{v=1}^m\|Z_v-P_vX\|_F^2 + \lambda\|P_v\|_F^2 \\
& = \min_{P_v}\|Z_v-P_vX\|_F^2 + \lambda\|P_v\|_F^2 \\
& = \min_{P_v}tr(Z_v-P_vX)^T(Z_v-P_vX) + \lambda tr(P_v^TP_v) \\
& = \min_{P_v}tr(-X^TP_v^TZ_v - Z_v^TP_vX + X^TP_v^TP_vX) + \lambda tr(P_v^TP_v)
\end{aligned}
\end{equation}

Thus, we take the derivatives of $\mathcal{L}$ w.r.t. $P_v$ and have:
\begin{equation} \label{equ24} 
\begin{aligned}
\nabla_{P_v}\mathcal{L} & = -\nabla_{P_v}tr{X^TP_v^TZ_v} - \nabla_{P_v}tr{Z_v^TP_vX} + \nabla_{P_v}tr{X^TP_v^TP_vX} + \lambda\nabla_{P_v}tr(P_v^TP_v) \\
& = -(XZ_v^T)^T - (XZ_v^T)^T + 2P_vXX^T + 2\lambda P_v\\
& = -2Z_vX^T + 2P_vXX^T + 2\lambda P_v \\
& = 0 
\end{aligned}
\end{equation}

Therefore, we obtain the close-form of $P_v$ as follows:
\begin{equation} \label{equ25} 
P_v = Z_vX^T(XX^T + \lambda\textbf{I})^{-1}
\end{equation}

\textbf{Next, we update $P_v^*$ by fixing the other parameters}, and thus the problem is reduced to the following formulation:
\begin{equation} \label{equ26} 
\min_{P_v^*}\mathcal{L} = \min_{P_v^*}\sum_{v=1}^m\|Z_v^*-P_v^*X^*\|_F^2 + \lambda^*\|P_v^*\|_F^2 \\
\end{equation}

Resembling solving for $P_v$, we derive the close-form solution of $P_v^*$ as follows:
\begin{equation} \label{equ27} 
P_v^* = Z_v^*{X^*}^T(X^*{X^*}^T + \lambda^*\textbf{I})^{-1}
\end{equation}

\textbf{Finally, we update $w, w^*, b, b^*$ by fixing the other parameters}, and thus the problem is reduced to solving for a classic SVM+ problem:
\begin{equation} \label{equ28} 
\begin{gathered} 
\min_{w, w^*, b, b^*}\mathcal{L} = \min_{w, w^*, b, b^*} \frac{1}{2}(\|w\|^2+\gamma\|w^*\|^2) + C\sum_{i=1}^n{w^*}^Tx_i^* + b^* \\
s.t.\quad y_i(w^Tx_i + b) \geq 1-(w^{*T}x_i^* + b^*), \quad w^{*T}x_i^* + b^* \geq 0,\quad \forall i=1,...,n 
\end{gathered}
\end{equation}
which can be solved by a fast algorithm in \cite{WenCVPR16}.

We iteratively alternate between the five steps until the objective function (\ref{equ8}) converges with global optimal solutions. The corresponding training process is summarized in Algorithm \ref{alg1}.

\begin{algorithm}[t]
\caption{Summary of our optimization procedure} \label{alg1} 
\begin{algorithmic}[1]
\REQUIRE\ $ \{z_v^{(i)}\}_{v=1}^m $, $ \{z_v^{*(i)}\}_{v=1}^m $, $ y_i $, $ \lambda $, $ \lambda^* $, $ \beta $, $ \beta^* $, $ \gamma $, $ C $, $ \quad \forall i=1,...,n $
\ENSURE\ $ \{x_i\}_{i=1}^n $, $ \{x_i^*\}_{i=1}^n $, $ \{P_v\}_{v=1}^m $, $ \{P_v^*\}_{v=1}^m $, $ w $, $ w^* $
\STATE Initialize: $ \{x_i^*\}_{i=1}^n $, $ \{P_v\}_{v=1}^m $, $ \{P_v^*\}_{v=1}^m $, $ w $, $ w^* $, $ b $, $ b^* $
\STATE \textbf{Repeat}
\STATE $\boldsymbol{\{x_i\}_{i=1}^n}$ \textbf{update} through solving Equ. (\ref{equ15}) by QP algorithm
\STATE $\boldsymbol{\{x_i^*\}_{i=1}^n}$ \textbf{update} through solving Equ. (\ref{equ20}) by QP algorithm
\STATE $\boldsymbol{\{P_v\}_{v=1}^m}$ \textbf{update} by Equ. (\ref{equ25})
\STATE $\boldsymbol{\{P_v^*\}_{v=1}^m}$ \textbf{update} by Equ. (\ref{equ27})
\STATE $\boldsymbol{w,w^*,b,b^*}$ \textbf{update} by SVM+ algorithm \cite{WenCVPR16}
\STATE \textbf{Until} Convergence
\end{algorithmic}
\end{algorithm}

In order to generate the latent representations $ \tilde{X} $ for on-the-fly re-ranking, we take the derivative of Equ. (\ref{equ11}) w.r.t. $ \tilde{X} $ and have:
\begin{equation} \label{equ29} 
\begin{aligned}
\nabla_{\tilde{X}}\mathcal{L}(\tilde{X}) & = \sum_{v=1}^m\nabla_{\tilde{X}}\|\tilde{Z}_v-\hat{P}_v\tilde{X}\|_F^2 + \beta\nabla_{\tilde{X}}\|\tilde{X}\|_F^2 \\
& = \sum_{v=1}^m\nabla_{\tilde{X}}tr(\tilde{Z}_v^T-\tilde{X}^T\hat{P}_v^T)(\tilde{Z}_v-\hat{P}_v\tilde{X}) + \beta\nabla_{\tilde{X}}tr(\tilde{X}^T\tilde{X}) \\
& = \sum_{v=1}^m\nabla_{\tilde{X}}tr(-\tilde{X}^T\hat{P}_v^T\tilde{Z}_v - \tilde{Z}_v^T\hat{P}_v\tilde{X} + \tilde{X}^T\hat{P}_v^T\hat{P}_v\tilde{X}) + \beta\nabla_{\tilde{X}}tr(\tilde{X}^T\tilde{X}) \\
& = \sum_{v=1}^m(-2\hat{P}_v^T\tilde{Z}_v + 2\hat{P}_v^T\hat{P}_v\tilde{X}) + 2\beta \tilde{X} \\
& = 0 \\
\end{aligned}
\end{equation}

Thus, we have the close-form solution of $ \tilde{X} $ as follows:
\begin{equation} \label{equ30} 
\tilde{X} = (\sum_{v=1}^m\hat{P}_v^T\hat{P}_v + \beta\textbf{I})^{-1}\sum_{v=1}^m\hat{P}_v^T\tilde{Z}_v
\end{equation}

\section{Analysis of the computational complexity} \label{section6}
We now discuss the computational complexity of our DMVPIR algorithm for separate phases. 

In the model training, the overall computational overhead consists of three main parts, i.e., solving for $ x $ and $ x^* $ in Equ. (\ref{equ15}) and (\ref{equ20}), computing $ P_v $ and $ P_v^* $ in Equ. (\ref{equ25}) and (\ref{equ27}) as well as updating $ w,w^*,b,b^* $ with SVM+ algorithm. Since both $ x $ and $ x^* $ are estimated by an off-the-shelf QP solver in practice, the corresponding time complexity can be computed as $ O(d^3) $, and thus the total cost for the whole training set in dual spaces accounts for $ 2n \cdot O(d^3) $. In Equ. (\ref{equ25}) and (\ref{equ27}), computing $ P_v $ and $ P_v^* $ requires $ O(D_v \cdot n \cdot d)+O(d^2 \cdot n)+O(d^3)+ O(D_v \cdot d^2)$ time complexity. It can be approximated by $ O(D_v \cdot n \cdot d)+O(D_v \cdot d^2 ) $, since $ D_v \gg d $ in our case. Thus, updating all the view-specific generation matrices in dual spaces takes $ 2\sum_{v=1}^mO(D_v \cdot n \cdot d)+O(D_v \cdot d^2 ) $. As for the $ w,w^*,b,b^* $ update, we directly use the fast linear SVM+ algorithm implemented in \cite{WenCVPR16}, and the time complexity is roughly $ O(2d)+O(d) $\cite{WenCVPR16}. Therefore, the total cost amounts to $ 2n \cdot O(d^3)+2\sum_{v=1}^mO(D_v \cdot n \cdot d)+O(D_v \cdot d^2)+O(2d)+O(d)$, which is thus reduced to $ 2n \cdot O(d^3)+2\sum_{v=1}^mO(D_v \cdot n \cdot d)+O(D_v \cdot d^2) $ approximately. 

During the re-ranking stage, the computational cost comprises the multi-view embedding for generating the latent representations shown in Equ. (\ref{equ30}) and the subsequent cosine similarity. The former is calculated as $ \sum_{v=1}^mO(d^2 \cdot D_v)+O(d^3)+\sum_{v=1}^mO(d \cdot D_v \cdot n)+O(d^2 \cdot n ) $ which can be approximated by $ \sum_{v=1}^mO(d^2 \cdot D_v)+O(d \cdot D_v \cdot n) $, while the latter accounts for $ O(n \cdot d) $ time complexity for efficient similarity measure.

\section{Experiments} \label{section7} 
In this section, we will evaluate our DMVPIR method for image re-ranking. First, we will introduce the public benchmark datasets as well as the experimental setup and the performance measure. Subsequently, thorough qualitative and quantitative evaluations will be carried out to demonstrate the performance of our approach. Besides, we also conduct a comparative study for showing the superiority of our method to the state-of-the-arts.

\subsection{benchmark datasets and performance measure}
We evaluate our DMVPIR re-ranking approach on two public datasets, Oxford5k \cite{PhilbinCVPR07} and Paris6k \cite{PhilbinCVPR08}, both of which are usually used as evaluation benchmarks for instance-level image retrieval. The two datasets include 5,063 and 6,392 images of 11 famous landmarks in Oxford and Paris respectively, and each landmark is represented by five query instances, which results in a total of 55 query groups used for querying the whole dataset. All the images in the dataset fall into four groups according to the query-specific relevance. Average Precision (AP) score is computed as the evaluation protocol for a single query, and thus mean Average Precision (mAP) is obtained by averaging all the AP scores for the overall performance measure. Besides, we also adopt the Normalized Discounted Cumulative Gain (NDCG) for evaluation \cite{JarvelinTOIS02}. The NDCG score at position $ P $ for a specific query can be computed as:
\begin{equation}
NDCG@P = Z_P\sum_{i=1}^P\frac{2^{l(i)} - 1}{log(i +1)}
\end{equation}
where $P$ is the ranking depth, $l(i)$ denotes the relevance of the $i^{th}$ ranked image to the specific query, and $Z_P$ is the normalization constant that makes the optimal NDCG@P equal 1. Similar to mAP, mean NDCG (mNDCG) score is also used for the overall performance evaluation.

\subsection{Multi-view features} \label{subsection 6.2}
Following \cite{JunTIP17}, we leverage three complementary image signatures for multi-view feature representations in our approach, namely \textbf{CNN}, \textbf{TE} and \textbf{VLAD+}. CNN feature is a 4,096-dimensional vector which consists in the activations of the upper layer of the deep VGG-16 architecture pretrained for the large-scale classification task \cite{SimonyanICLR15}. Known as the best shallow image signature thus far, TE referred to as triangulation embedding is viewed as a promising alternative to FV vector \cite{JegouCVPR14}, whilst VLAD+ developed from RootSIFT descriptor is more computationally efficient for fast retrieval \cite{ArandjelovicCVPR13}. In implementation, we use the same vocabulary sizes for TE and VLAD+ as in \cite{JunTIP17}, which leads to 8,064 and 16,384-dimensional vectors for respective representations. The complementarity among the three heterogeneous features can be fully exploited for multiple feature embedding, since deep CNN feature enables high-level image description, whilst TE and VLAD+ inherit desirable invariant property from robust local descriptors.

\subsection{Interactive relevance feedback with PI annotation}
Analogous to \cite{JunTIP17}, given the ranking images obtained in the first place, we utilize the user relevance feedback (URF) performed once for assembling the positive query-relevant images while automatically recognize the low-scored examples as the negative distractors for training our re-ranking model. Different from the conventional URF methods \cite{TaoPAMI06,JunTIP17,WangNEUCOM14}, however, not only a click indicating the query-relevance of an image but also the object ROI capturing the user query is required for obtaining the auxiliary PI data our scenario. To be specific, we annotate the image ROIs in the positive examples while adopt the off-the-shelf saliency detector \cite{FengICCV11} for generating the PI regions in the negative images. Thus, the original set of training images alongside the corresponding supplementary PI data are delivered to the subsequent module for extracting multi-view features. Since the user interaction with PI annotation is performed on the shortlisted images relatively accounting for a small proportion of the top returned results, this practice incurs affordable overhead on the system.

\subsection{Model selection}
In DMVPIR, six hyperparamters in Equ. (\ref{equ8}) need to be carefully tuned, i.e., $ \lambda $, $ \lambda^* $, $ \beta $, $ \beta^* $, $ \gamma $, $ C $. To this end, we perform model selection on a single query, and the optimal parameters obtained accordingly are used for evaluating the other query groups on the two benchmark datasets. In implementation, we select the query ``all\_souls\_1'' for model training with varying parameters.

\subsection{Experimental results}
\subsubsection{Comparison of baseline methods}
In our baseline retrieval systems, a global image signature is combined with efficient cosine similarity for generating a set of ranking images in the first place. In our case, we evaluate three image representations introduced in section \ref{subsection 6.2}, which leads to different baseline methods respectively denoted as \textbf{TE\_cos}, \textbf{CNN\_cos} and \textbf{VLAD+\_cos}. Table \ref{tab2} gives the performance of different baselines. It is clearly shown that TE\_cos consistently outperforms the other two approaches by achieving highest mAP at 61.76\% and 62.04\% on the respective datasets as well as higher mNDCG scores. Surprisingly, CNN\_cos exhibits the suboptimal performance inferior to TE\_cos, which can be attributed to the pre-trained deep model with insufficient descriptive power. Although fine-tuning allows further improving the retrieval performance of CNN\_cos, we still use the TE\_cos as the baseline for the subsequent re-ranking, since in our work we only focus on the image re-ranking which operates independently of the baseline method.

\begin{table} \addtolength{\tabcolsep}{-0.1cm} \renewcommand{\arraystretch}{1.15}
\centering
\caption{Comparison of the three baseline methods on Oxford5k and Paris6k (\%).}
\vspace{.1cm}
\label{tab2} 
\begin{tabular}{|c|c|c|c|c|c|c|}
\hline 
\multirow{2}{*}{\shortstack{Performance\\Measure}} & \multicolumn{3}{|c|}{Paris6k} & \multicolumn{3}{|c|}{Oxford5k} \\ \cline{2-7}
 & TE\_cos & CNN\_cos & VLAD+\_cos & TE\_cos & CNN\_cos & VLAD+\_cos \\ 
\hline 
mAP & \textbf{61.76} & 58.75 & 49.15 & \textbf{62.04} & 45.05 & 46.98 \\ 
\hline 
mNDCG@50 & \textbf{87.70} & 83.99 & 80.02 & \textbf{70.27} & 59.88 & 59.71 \\ 
\hline 
mNDCG@100 & \textbf{77.96} & 74.23 & 67.50 & \textbf{70.16} & 60.49 & 59.47 \\ 
\hline 
\end{tabular}
\end{table}

\subsubsection{The performance of our DMVPIR method}
\begin{figure} 
\centering
\includegraphics[width=1.0\linewidth]{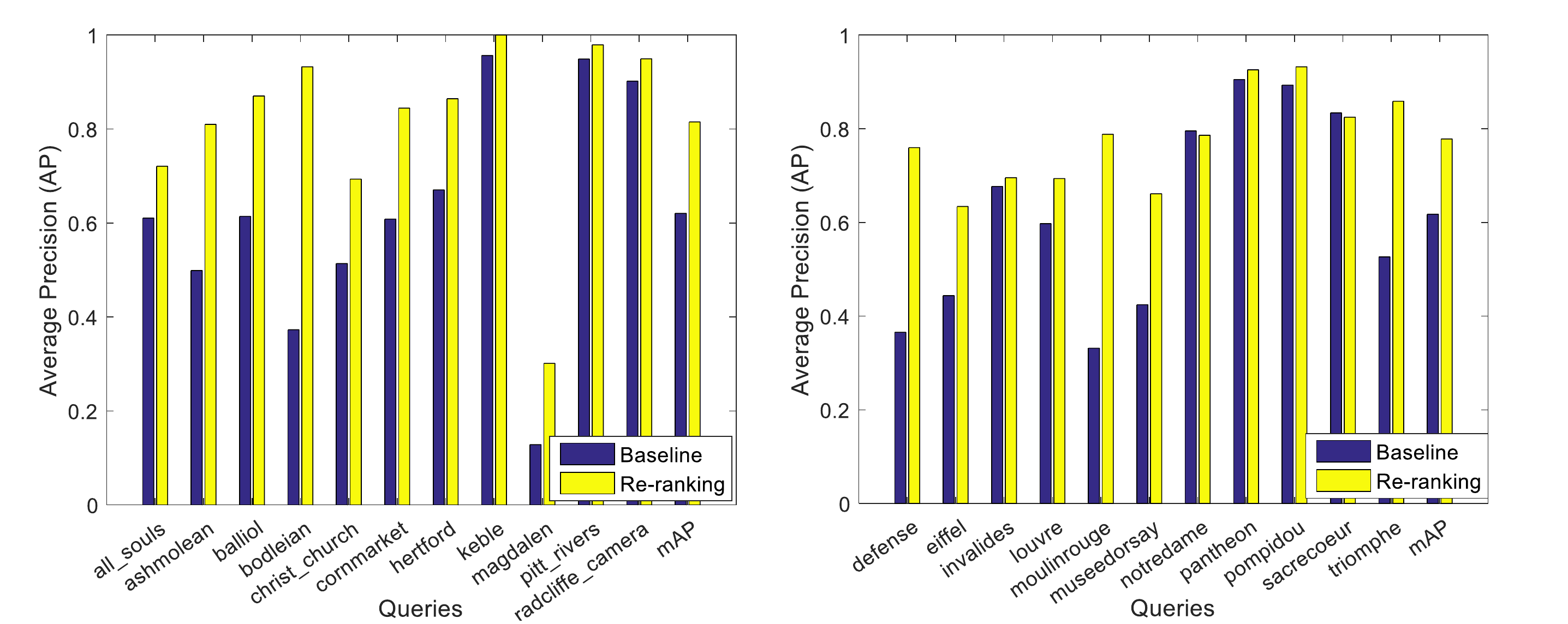}
\caption{Comparison of the baseline and our re-ranking method on Oxford5k (left) and Paris6k (right) in terms of AP score.} \label{fig3} 
\end{figure}

We impose our DMVPIR method on the baseline TE\_cos for accurate re-ranking. Fig. \ref{fig3} presents the comparison of the baseline and our re-ranking approach in terms of AP score. It is observed that DMVPIR provides significant performance gains ranging from 1.9\% on ``invalides'' to 56\% on ``bodleian'' for different query groups. In particular, DMVPIR reports respective mAP scores at \textbf{81.51\%} and \textbf{77.83\%} on two datasets and outperforms the baseline system by approximately \textbf{20\%} and \textbf{16\%}, which substantially suggests the beneficial effect of the proposed re-ranking approach. The only exceptions come from the queries ``notredame'' and ``sacrecoeur'' when slight performance drop occurs. This implies the generalization capability of DMVPIR is somewhat prone to the high nonlinearity of our model and the redundancy occasionally present in the training examples.

In addition, we compare the baseline and DMVPIR methods by computing NDCG scores. As shown in Table \ref{tab3} and \ref{tab4}, DMVPIR dramatically boosts the baseline results from 70.27\% to \textbf{79.30\%} on Oxford5k while the performance gains also reach 7\% on Paris 6k in terms of mNDCG@50. Similar trend can also be observed for mNDCG@100 score on both datasets, which demonstrates our re-ranking method considerably benefits the performance improvement.

\begin{table} \addtolength{\tabcolsep}{0.0cm} \renewcommand{\arraystretch}{1.15}
\centering
\caption{Comparison of baseline and DMVPIR with NDCG scores on Oxford5k.} \label{tab3}
\vspace{+.1cm}
\begin{tabular}{|c|c|c|c|c|}
\hline 
\multirow{2}{*}{Query} & \multicolumn{2}{|c|}{\textbf{Baseline}} & \multicolumn{2}{|c|}{\textbf{Re-ranking}} \\ \cline{2-5} 
 & NDCG@50 & NDCG@100 & NDCG@50 & NDCG@100 \\ 
\hline
\hline 
all\_souls & 0.7206 & 0.6683 & 0.7261 & 0.6843 \\ 
\hline 
ashmolean & 0.6617 & 0.6825 & 0.7858 & 0.7954 \\ 
\hline 
balliol & 0.6561 & 0.6860 & 0.7220 & 0.7305 \\ 
\hline 
bodleian & 0.5847 & 0.6297 & 0.9030 & 0.9133 \\ 
\hline 
christ\_church & 0.6259 & 0.6033 & 0.8136 & 0.7482 \\ 
\hline 
cornmarket & 0.8137 & 0.8137 & 0.9188 & 0.9252 \\ 
\hline 
hertford & 0.7462 & 0.7740 & 0.8318 & 0.8288 \\ 
\hline 
keble & 0.9486 & 0.9511 & 0.8775 & 0.8775 \\ 
\hline 
magdalen & 0.2492 & 0.2385 & 0.4471 & 0.4148 \\ 
\hline 
pitt\_rivers & 0.8830 & 0.8902 & 0.9161 & 0.9161 \\ 
\hline 
radcliffe\_camera & 0.8400 & 0.7803 & 0.7806 & 0.7656 \\ 
\hline 
mean & 0.7027 & 0.7016 & \textbf{0.7930} & \textbf{0.7818} \\ 
\hline 
\end{tabular} 
\end{table}

\begin{table} \addtolength{\tabcolsep}{0.0cm} \renewcommand{\arraystretch}{1.15}
\centering
\caption{Comparison of baseline and DMVPIR with NDCG scores on Paris6k.} \label{tab4}
\vspace{+.1cm}
\begin{tabular}{|c|c|c|c|c|}
\hline 
\multirow{2}{*}{Query} & \multicolumn{2}{|c|}{\textbf{Baseline}} & \multicolumn{2}{|c|}{\textbf{Re-ranking}} \\ \cline{2-5} 
 & NDCG@50 & NDCG@100 & NDCG@50 & NDCG@100 \\ 
\hline
\hline 
defense & 0.7482 & 0.5309 & 0.9592 & 0.7936 \\ 
\hline 
eiffel & 0.8872 & 0.8002 & 0.9260 & 0.8643 \\ 
\hline 
invalides & 0.9852 & 0.9378 & 0.9884 & 0.9473 \\ 
\hline 
louvre & 0.8224 & 0.7279 & 0.8674 & 0.7566 \\ 
\hline 
moulinrouge & 0.7690 & 0.6163 & 0.9882 & 0.9627 \\ 
\hline 
museedorsay & 0.6485 & 0.5262 & 0.8407 & 0.6952 \\ 
\hline 
notredame & 0.9910 & 0.8956 & 0.9860 & 0.8883 \\ 
\hline 
pantheon & 0.9941 & 0.9355 & 0.9982 & 0.9559 \\ 
\hline 
pompidou & 0.8641 & 0.8234 & 0.9008 & 0.8874 \\ 
\hline 
sacrecoeur & 0.9647 & 0.9084 & 0.9780 & 0.9257 \\ 
\hline 
triomphe & 0.9727 & 0.8734 & 0.9907 & 0.9745 \\ 
\hline 
mean & 0.8770 & 0.7796 & \textbf{0.9476} & \textbf{0.8774} \\ 
\hline 
\end{tabular} 
\end{table}
 
\subsubsection{The comparative studies} 
In comparative studies, we compare our approach DMVPIR with other multi-view re-ranking methods as follows:

1) DMINTIR. We directly reproduce the algorithm in \cite{JunTIP17} with the analogous parameter setting adopted in our method.

2) DMINTIR-PI. For this approach, we leverage the local multi-view PI features for learning the separating hyperplane $w$ without taking into account the original multiple global feature representations. The online re-ranking is achieved by computing and sorting the distances from the global multi-view projections of the target images to the hyperplane $w$.

3) DQE by Concatenating Averaged Reduced-size Multi-View features for Re-ranking (DQE-CAR-MVR). We first impose PCA on the multi-view features for dimension reduction in both original and privileged space. Thus, we fuse the compressed view-specific features in the two spaces by average pooling and concatenate the pooled features of different views for the holistic representation. Subsequently, analogous to \cite{ArandjelovicCVPR12}, we train a linear SVM model on the resulting representation and compute the signed distance from the separating hyperplane for re-ranking. Note that the reduced feature dimensionality in this method is also set to be 128, which is consistent with the setting in our approach.

4) DQE by Concatenating Averaged Full-size Multi-View features for Re-ranking (DQE-CAF-MVR). This method is essentially the same with DQE-CAR-MVR except the original dimensionalities of the multi-view features are maintained without dimension reduction.

5) DQE by Averaging Reduced-size Multi-view features for Re-ranking (DQE-AR-MVR). Different from DQE-CAR-MVP and DQE-CAF-MVP, this approach directly utilizes average pooling for fusing all the multi-view features with reduced size in both spaces, which leads to the final image representation delivered to the linear SVM model. The reduced feature size is also set to be 128 for the sake of consistency.

6) Late Fusion on DQE with Averaged Reduced-size Multi-view features for Re-ranking (LFDQE-AR-MVR). In this method, the size of the multi-view features in dual spaces are firstly reduced by PCA and view-specific average pooling is also performed for generating fused representation similar to DQE-CAR-MVR. Then, we derive multiple DQE models from respective fused features and combine the output for the relevance score in re-ranking.

7) Late Fusion on DQE with Averaged Full-size Multi-view features for Re-ranking (LFDQE-AF-MVR). Different from LFDQE-AR-MVR, this approach adopts the full size of the multiple features without dimension reduction for respective DQE model training.

To sum up, both DMINTIR and DMINTIR-PI simply take into account the visual information in a single space, whilst our approach along with the other competing methods combine the visual contents from both spaces. In particular, DQE-CAR-MVR, DQE-CAF-MVR as well as DQE-AR-MVR can be viewed as early fusion multi-view re-ranking strategies, whilst LFDQE-AR-MVR and LFDQE-AF-MVR fall into the category of late fusion techniques.

Table. \ref{tab5} and \ref{tab6} present the performance of different multi-view re-ranking methods on the two benchmarks. Overall, our scheme demonstrates the unrivalled performance superior to the other competing approaches. In particular, the proposed method performs better than both DMINTIR and DMINTIR-PI, which implies the considerable benefit in combining the original visual clues with supplementary PI data for re-ranking. More specifically, DMVPIR reports higher mAP scores surpassing DMINTIR by 1.2\% and 0.7\% respectively on two datasets. Since there exists the asymmetry between the training and the testing information in DMINTIR-PI, DMVPIR exhibits more dramatic performance advantage against DMINTIR-PI with significant improvements over 15\%. In addition, our method also beats the other fusion-based re-ranking method by achieving substantial performance gains. This sufficiently suggests our subspace-based scheme allows learning the discriminative representation from heterogeneous multi-view features while works better then the methods which perform straightforward fusion strategies. Note that our scheme does not achieve the best mNDCG results on Oxford5k. We argue this results from the evaluation mechanism of NDCG where the junk images with certain ambiguity are also taken into consideration in computing the query-relevance, whereas they are discarded in evaluating mAP score. In this sense, our scheme enables having clear groundtruth images returned at higher ranks than those ambiguous examples.

\begin{table} \addtolength{\tabcolsep}{0.1cm} \renewcommand{\arraystretch}{1.2} 
\centering 
\caption{Comparison of different classification-based multi-view re-ranking methods on Oxford5k (\%).}
\vspace{.1cm}
\label{tab5}
\begin{tabular}{|c|c|c|c|}
\hline 
Methods & mAP & mNDCG@50 & mNDCG@100 \\ 
\hline
\hline 
DMINTIR & 80.34 & \textbf{82.82} & \textbf{81.13} \\ 
\hline 
DMINTIR-PI & 61.56 & 66.83 & 66.72 \\ 
\hline 
DQE-CAR-MVR & 77.39 & 79.23 & 77.91 \\ 
\hline 
DQE-CAF-MVR & 78.72 & 79.72 & 78.01 \\ 
\hline 
DQE-AR-MVR & 40.37 & 49.47 & 49.36 \\
\hline
LFDQE-AR-MVR & 74.42 & 77.43 & 75.80 \\ 
\hline
LFDQE-AF-MVR & 79.48 & 80.89 & 78.89 \\
\hline
\hline 
Ours & \textbf{81.51} & 79.30 & 78.18\\ 
\hline 
\end{tabular}
\end{table} 

\begin{table} \addtolength{\tabcolsep}{0.1cm} \renewcommand{\arraystretch}{1.2} 
\centering 
\caption{Comparison of different classification-based multi-view re-ranking methods on Paris6k (\%).}
\vspace{.1cm}
\label{tab6}
\begin{tabular}{|c|c|c|c|}
\hline 
Methods & mAP & mNDCG@50 & mNDCG@100 \\ 
\hline
\hline 
DMINTIR & 77.09 & 94.64 & 87.28 \\ 
\hline 
DMINTIR-PI & 61.91 & 85.88 & 76.22 \\ 
\hline 
DQE-CAR-MVR & 72.36 & 92.22 & 83.85 \\ 
\hline 
DQE-CAF-MVR & 74.90 & 93.85 & 85.13 \\ 
\hline 
DQE-AR-MVR & 46.89 & 63.70 & 58.04 \\
\hline
LFDQE-AR-MVR & 64.54 & 83.00 & 75.69 \\ 
\hline
LFDQE-AF-MVR & 74.57 & 93.80 & 85.23 \\ 
\hline
\hline 
Ours & \textbf{77.83} & \textbf{94.76} & \textbf{87.74} \\ 
\hline 
\end{tabular} 
\end{table}

Besides, we also compare the proposed DMVPIR method with the state-of-the-arts in recent years. As illustrated in Table \ref{tab7}, DMVPIR achieves performance on par with both traditional BoW-based and recent CNN-based re-ranking approaches, which substantially suggests the promise of the proposed framework. In particular, compared with CNN-based methods, our approach significantly surpasses \cite{KalantidisECCVW16} on Oxford5k by over 14\% and reports comparable result on Paris6k with the same feature size. Additionally, DMVPIR consistently beats Faster R-CNN+CA-SR+QE which also makes use of the deep model pre-trained with VGG16 architecture \cite{SimonyanICLR15} while enjoys a more compact representation. Although fine-tuning the VGG16 network brings further performance gains, DMVPIR still achieves higher re-ranking accuracy than Faster R-CNN+CS-SR+QE on Oxford5k and rivals the performance on Paris6k. 

\begin{threeparttable} \addtolength{\tabcolsep}{0.3cm} \renewcommand{\arraystretch}{1.2}
\centering
\caption{Comparison of our approach and the state-of-the-art re-ranking methods on two datasets(mAP). $ d $ and $ K $ refers to the feature dimensionality and the vocabulary size respectively.}
\label{tab7} 
\vspace{.1cm}
\begin{tabular}{|c|c|c|c|}
\hline
BoW-based Methods & $ K $ & Oxford5k & Paris6k \\ 
\hline 
Recoprocal NN \cite{QinCVPR11} & 500k & 81.4 & 80.3 \\
\hline
Database Saliency \cite{GaoTMM15} & 1024 & 0.835 & 0.814 \\ 
\hline
HE+MA+PGM \cite{LiCVPR15} & 100k & 0.737 & - \\
\hline
LS+R+LQE \cite{MohedanoICMR16} & 25k & 0.788 & 0.848 \\
\hline
\hline
CNN-based Methods & $ d $ & Oxford5k & Paris6k \\
\hline
\multirow{3}{*}{CroW + QE \cite{KalantidisECCVW16}} & 128 & 0.670 & 0.793 \\ \cline{2-4}
 & 256 & 0.718 & 0.815 \\ \cline{2-4}
 & 512 & 0.749 & 0.848 \\
\hline
R-MAC+AML+QE \cite{ToliasICLR16} & 512 & 0.773 & 0.865 \\
\hline
Faster R-CNN+CA-SR+QE \cite{SalvadorCVPRW16} & 512 & 0.647 & 0.732 \\
\hline
\multirow{2}{*}{Faster R-CNN+CS-SR+QE$ ^\star $\cite{SalvadorCVPRW16}} & 512 & 0.678 & 0.784 \\ \cline{2-4}
 & 512 & 0.786 & 0.842 \\
\hline
\hline
Ours & 128 & \textbf{0.8151} & \textbf{0.7783} \\
\hline
\end{tabular}
\begin{tablenotes}
\small
\item $ ^{\star} $achieved with two different fine-tuning strategies
\end{tablenotes}
\vspace{.3cm}
\end{threeparttable}

\begin{figure}
\centering
\includegraphics[width=1.0\linewidth]{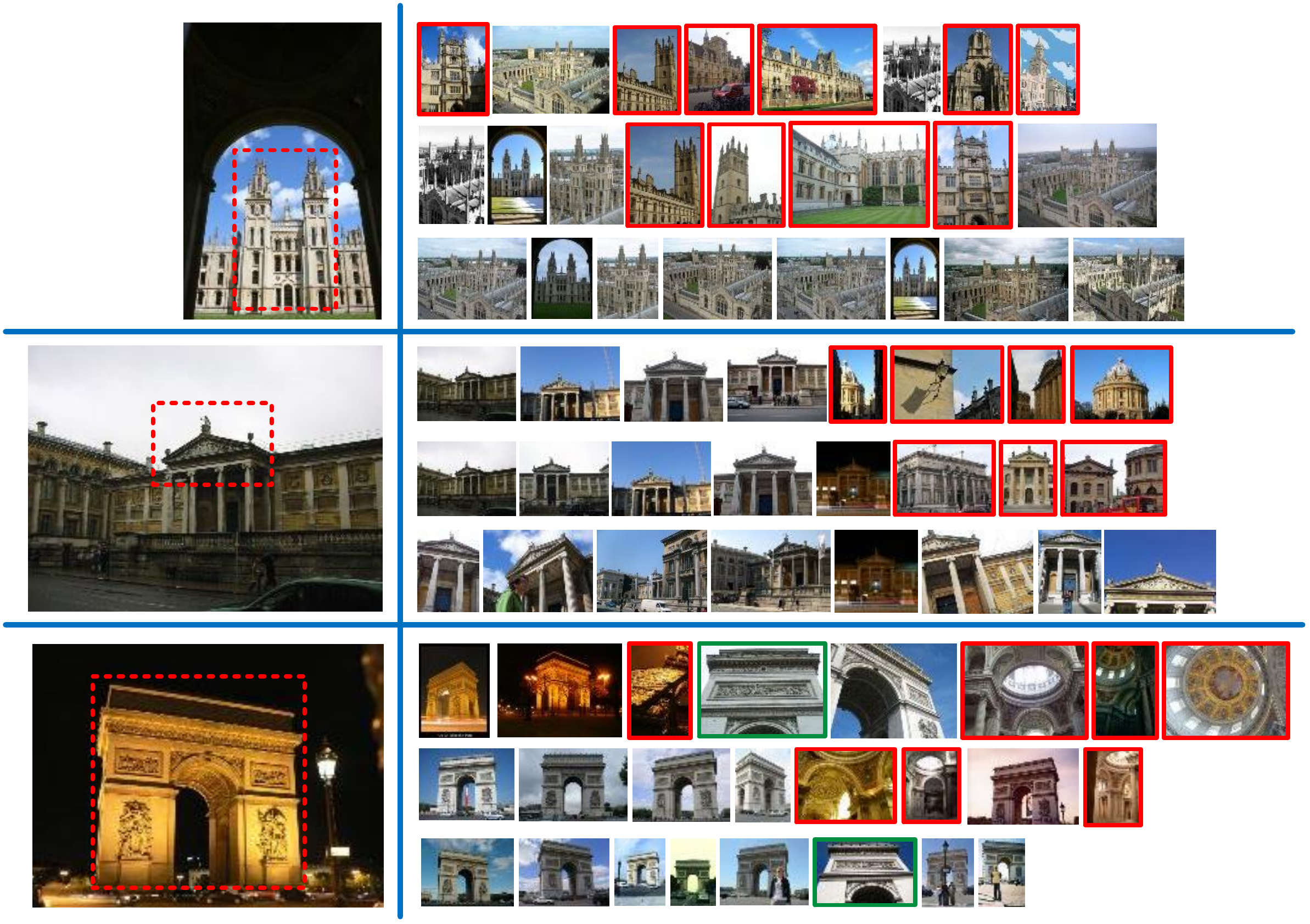}
\caption{Comparison of qualitative retrieval results achieved by baseline (the first row), DMINTIR \cite{JunTIP17} (the second row) and our scheme (the last row). Given an image with the annotated query region outlined by the red dashed box shown on the left, the top returned results are displayed accordingly on the right. Note that the junk images and the false alarms are highlighted in green and red boxes, respectively. It is observed that our approach significantly improves the baseline result and exhibits performance advantage against DMINTIR by returning more top ranked query-related images. }
\label{fig4} 
\end{figure}

In addition to the above quantitative evaluations, we also present the qualitative results of different methods as shown in Fig. \ref{fig4}. It is observed that our scheme not only significantly improves the retrieval accuracy of the baseline but also demonstrates better performance than the-state-of-the-art DMINTIR method. Specifically, with the help of PI learning, our approach enables returning more top ranked ground-truth images even when the query-related instances only account for small regions with the surrounding complex visual background or are partially occluded by other objects (e.g., tree, person, lamp post) in the image. This sufficiently suggests incorporating PI learning in re-ranking contributes to further improving the retrieval performance.

\subsubsection{Quantitative computational cost}
We quantitatively evaluate the time cost of the proposed DMVPIR algorithm in separate steps, i.e., the model training and on-the-fly re-ranking. As illustrated in table \ref{tab8}, training query-specific model costs approximately \textbf{80s} while fast on-the-fly re-ranking can be achieved with not more than \textbf{0.2s}, which indicates that our scheme lends itself to the real-time scenarios. In practice, all the experiments are conducted using Matlab on a machine with 3.20GHz Intel Core i5-6500 CPU and 8GB memory.

\begin{table} \addtolength{\tabcolsep}{0.2cm} \renewcommand{\arraystretch}{1.2}
\centering
\caption{Time cost of our scheme on two datasets(s).}
\label{tab8} 
\vspace{.1cm}
\begin{tabular}{|c|c|c|}
\hline 
Datasets & model training & re-ranking \\ 
\hline 
Oxford5k & 79.35 & 0.14 \\ 
\hline 
Paris6k & 78.66 & 0.18 \\ 
\hline 
\end{tabular} 
\end{table}

\subsection{Parameter analysis}
We now thoroughly discuss the impact of various parameters in the proposed DMVPIR framework on the re-ranking performance, including the vocabulary size $ k $ for generating the TE and VLAD+ features, the six hyperparameters to tune in Equ. (\ref{equ8}), the length of the shortlist $ K $ for user interaction and the subspace dimensionality $ d $.

\subsubsection{The impact of the vocabulary size}
In our method, the re-ranking performance largely depends on the multiple features including TE and VLAD+ both of which need a well-trained vocabulary. As for the TE signature, we follow the standard practice \cite{JegouCVPR14,BabenkoICCV15,TaoCVPR15} to set the vocabulary size $ k $ as 64 for generating a 8,064 dimensional feature with low frequency dimensions removed, since further increasing $ k $ yields limited boost in performance while severely compromises the computational efficiency \cite{JegouCVPR14}. In terms of VLAD+, we use the vocabulary of the same size as in \cite{JunTIP17}. In order to explore the impact of $ k $ on the performance, we further increase $ k $ to 256 and 1024 respectively. Consistent with \cite{JunTIP17}, the resulting performance gains consist in less than 1\% and 1.5\% at the cost of considerable growth in memory footprint and computational overhead. Therefore, we use $ k=256 $ for VLAD+ in all tests for the tradeoff between accuracy and efficiency.

\subsubsection{The impact of tradeoff hyperparameters}
\begin{figure}
\centering
\includegraphics[width=1.0\linewidth]{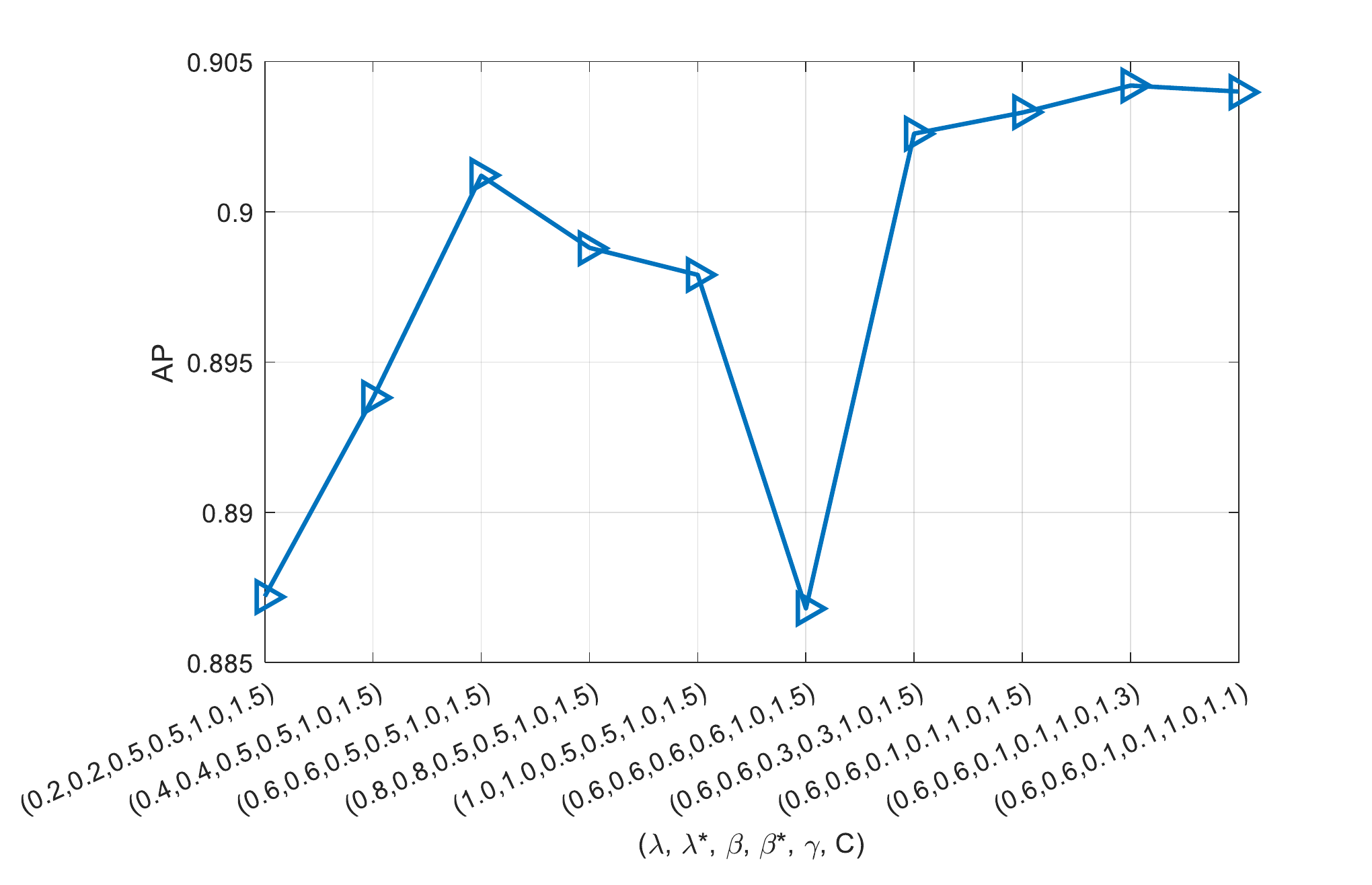}
\caption{The parameter analysis on query group ``all\_souls\_1''.}
\label{fig5} 
\end{figure}

For model selection, we evaluate different combinations of hyperparameters $ \{\lambda,\lambda^*,\beta,\beta^*,\gamma, C\} $ on a single query group ``all\_souls\_1'' to obtain the optimal ones. As illustrated in Fig. \ref{fig5}, the highest AP score is achieved at 90.42\% when the hyperparameters take the values of $ \{0.6,0.6,0.1,0.1,1.0,1.3\} $. Thus, we use the set $ \{0.6,0.6,0.1,0.1,1.0,1.3\} $ for evaluations on both datasets. Overall, the performance with different hyperparameter combinations fluctuates slightly from 88.72\% to 90.42\%, which, to some extent, implies the desirable property of DMVPIR in hyperparameter insensitivity. This can be explained by the fact that introducing PI into our framework brings the performance boost varying within a certain range dependent on the tradeoff between respective regularization terms.

\subsubsection{The influence of the user interaction}
\begin{figure}
\centering
\includegraphics[width=1.0\textwidth]{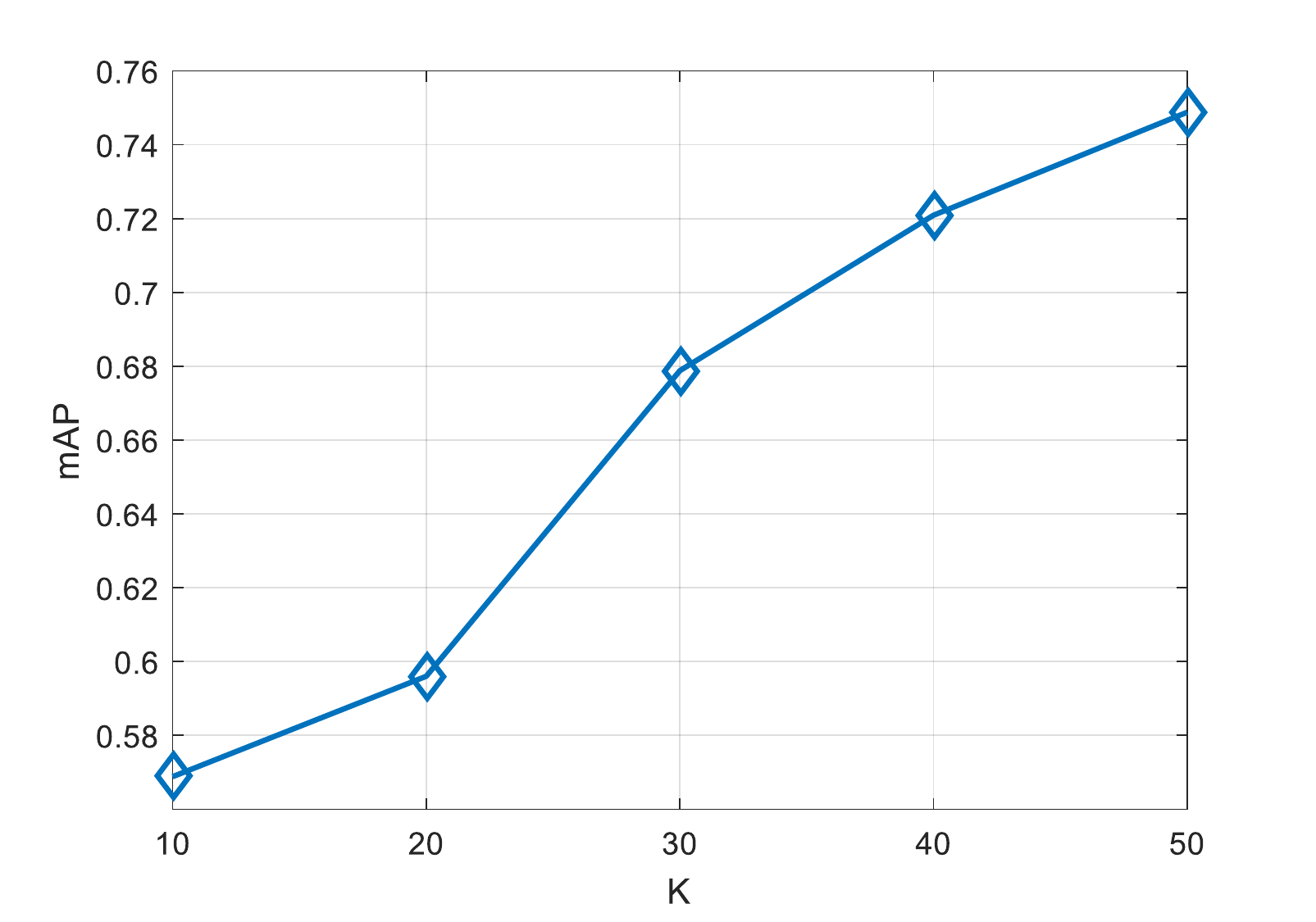}
\caption{The effect of the user interaction on the query ``all\_souls''.}
\label{fig6} 
\end{figure}

Analogous to \cite{JunTIP17}, user interaction is involved in training DMVPIR model for obtaining the query-relevant positive images with annotated PI regions from the top returned shortlist. Thus, it is essential to explore the effect of the shortlist size $ K $ on the re-ranking performance. Fig. \ref{fig6} gives the DMVPIR performance with varying $ K $ on the query ``all\_souls''. It is shown that the re-ranking accuracy improves with an increase in $ K $, yet the growth declines, which implies further increasing $ K $ leads to limited performance improvements at the cost of more user interaction and human workload. In practise, we do not take into account the case when $ K $ takes the value greater than 50, since not only a user click indicating the query-relevance but also annotating the PI region is required in our case. Therefore, larger $ K $ tends to incur unaffordable burden and thus adversely affect the efficiency of the whole system. In implementation, we assume $ K=40 $ is a reasonable choice with desirable compromise between accuracy and efficiency. Since the images with low ranks are recognized as the negative training data without using any user interaction, the size of the negative set is empirically set to be 100. Thus, we use this parameter setting (40/100) for all query groups.

\subsubsection{The effect of the subspace dimension}
\begin{figure}
\centering
\includegraphics[width=1.0\textwidth]{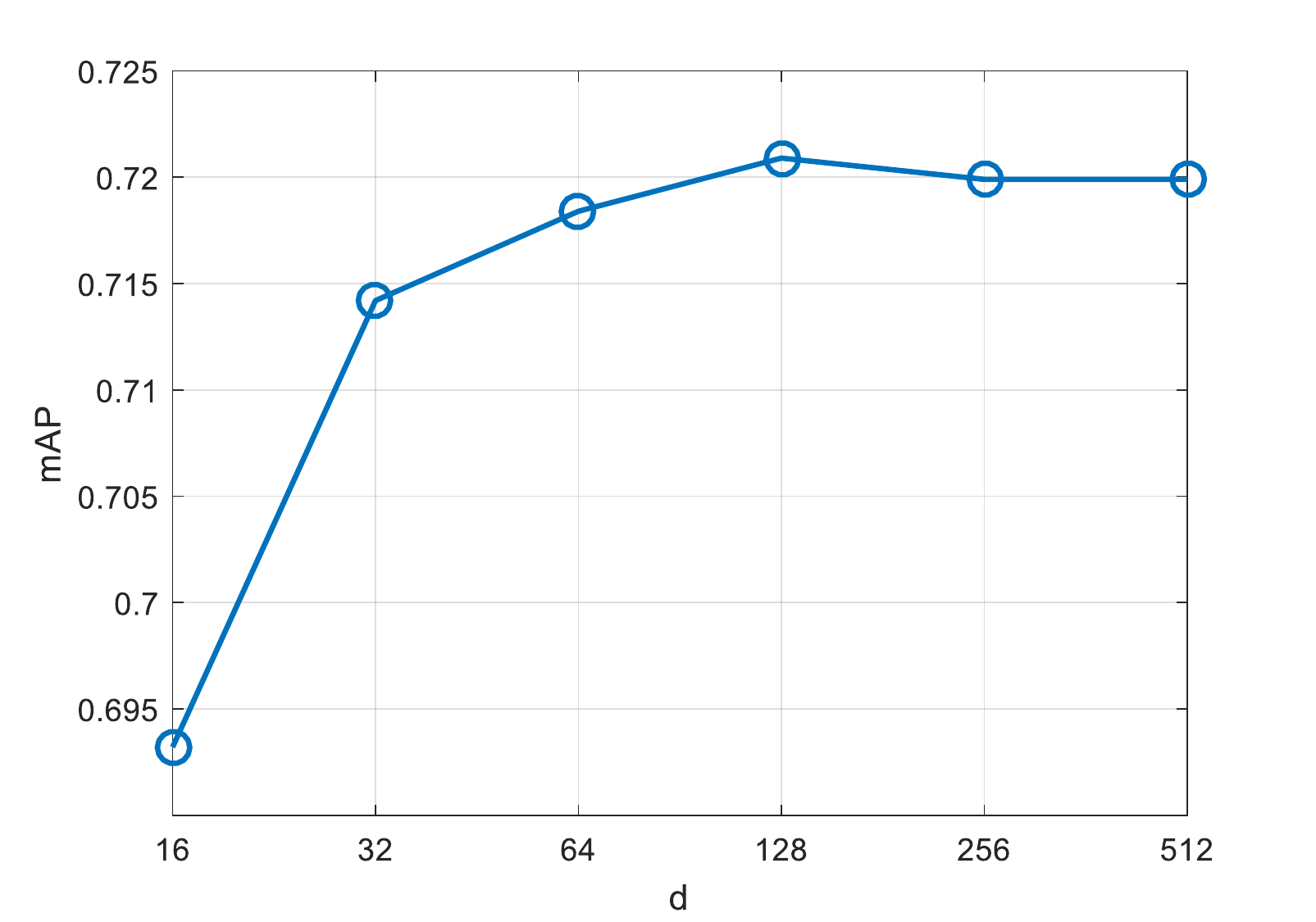}
\caption{The performance of DMVPIR with different subspace dimensions $ d $.}
\label{fig7} 
\end{figure}

Fig. \ref{fig7} illustrates the performance of our approach with different low-dimensional subspaces on query ``all\_souls''. Overall, the retrieval performance grows with an increase in $ d $ when the highest mAP score is reported at 72.09\% with $ d=128 $. Besides, a slight performance drop is observed when the subspace dimension exceeds 128. Interestingly, increasing the subspace dimension does not bring further performance boost, which sufficiently implies the feature redundancy present in the original multi-view spaces. As a result, we use the 128-dimensional subspace in our scenario.

\section{Conclusion and future work} \label{section8}
In this paper, we propose a discriminative multi-view PI-aware image re-ranking method termed as DMINTIR. Different from the conventional multi-view re-ranking approaches, we take into consideration the supplementary PI cues, since they are capable of characterising the dominant information in the image that captures the query intention. In model training, the auxiliary PI data and the original training data are simultaneously delivered to the unified multi-view embedding framework for producing a PI-aware subspace with sufficient discriminating power. For accurate re-ranking, the PI-aware latent representations can be obtained by projecting the multi-view features of the target images onto the underlying space for efficient similarity measure. Extensive evaluations on the public datasets for landmark retrieval task demonstrate our scheme outperforms the classical multi-view re-ranking strategies and achieves the comparable results on par with the state-of-the-arts.

Despite effective, DMINTIR somewhat relies on the user interaction for PI annotation. In the future, we will further study the generalization capability of the re-ranking model when the PI cues are limited. Besides, improving the efficiency and the scalability of our algorithm is another line of research in our future work.


%


\end{document}